# Chapter 5.

# DISENTANGLEMENT FOR DISCRIMINATIVE VISUAL RECOGNITION

## Xiaofeng Liu

*Harvard University, Beth Israel Deaconess Medical Center, Harvard Medical School, Boston, MA, USA*
*xliu11@bidmc.harvard.edu*

**Annotation**

Recent successes of deep learning-based recognition rely on maintaining the content related to the main-task label. However, how to explicitly dispel the noisy signals for better generalization in a controllable manner remains an open issue. For instance, various factors such as identity-specific attributes, pose, illumination and expression affect the appearance of face images. Disentangling the identity-specific factors is potentially beneficial for facial expression recognition (FER). This chapter systematically summarize the detrimental factors as task-relevant/irrelevant semantic variations and unspecified latent variation. In this chapter, these problems are casted as either a deep metric learning problem or an adversarial minimax game in the latent space. For the former choice, a generalized adaptive (N+M)-tuplet clusters loss function together with the identity-aware hard-negative mining and online positive mining scheme can be used for identity-invariant FER. The better FER performance can be achieved by combining the deep metric loss and softmax loss in a unified two fully connected layer branches framework via joint optimization. For the latter solution, it is possible to equipping an end-to-end conditional adversarial network with the ability to decompose an input sample into three complementary parts. The discriminative representation inherits the desired invariance property guided by prior knowledge of the task, which is marginal independent to the task-relevant/irrelevant semantic and latent variations. The framework achieves top performance on a serial of tasks, including lighting, makeup, disguise-tolerant face recognition and facial attributes recognition. This chapter systematically summarize the popular and practical solution for disentanglement to achieve more discriminative visual recognition.

## 5.1. Introduction

Extracting a discriminative representation for the task at hand is an important research goal of recognition [Liu, et al., 2019], [Liu, et al., 2019], [Liu, et al., 2018]. The typical deep learning solution utilize the cross-entropy loss to enforce the extracted feature representation has the sufficient information about the label [Liu, et al., 2019]. However, this setting does not require the extracted representation is purely focus on the label, and usually incorporate the unnecessary information that not related to the label [Liu, et al., 2019].

143

For example, the identity information in the facial expression recognition feature. Then, the perturbation of the identity will unavoidable result the change of expression feature [Liu, et al., 2017]. These identity-specific factors degrade the FER performance of new identities unseen in the training data [Liu, et al., 2019]. Since spontaneous expressions only involve subtle facial muscle movements, the extracted expression-related information from different classes can be dominated by the sharp-contrast identity-specific geometric or appearance features which are not useful for FER. As shown in Fig.5.1, example $x_1$ and $x_3$ are of happy faces whereas $x_2$ and $x_4$ are not of happy faces. $f(x_i)$ are the image representations using the extracted features. For FER, it is desired that two face images with the same expression label are close to each other in the feature space, while face images with different expressions are farther apart from each other, i.e., the distance $D_2$ between examples $x_1$ and $x_3$ should be smaller than $D_1$ and $D_3$, as in Fig.5.1 (b). However, the learned expression representations may contain irrelevant identity information as illustrated in Fig.5.1 (a). Due to large inter-identity variations, $D_2$ usually has a large value while the $D_1$ and $D_3$ are relatively small. Similarly, the expression related factors will also affect the recognition of face identity. Actually, the pure feature representation is the guarantee of a robust recognition system.

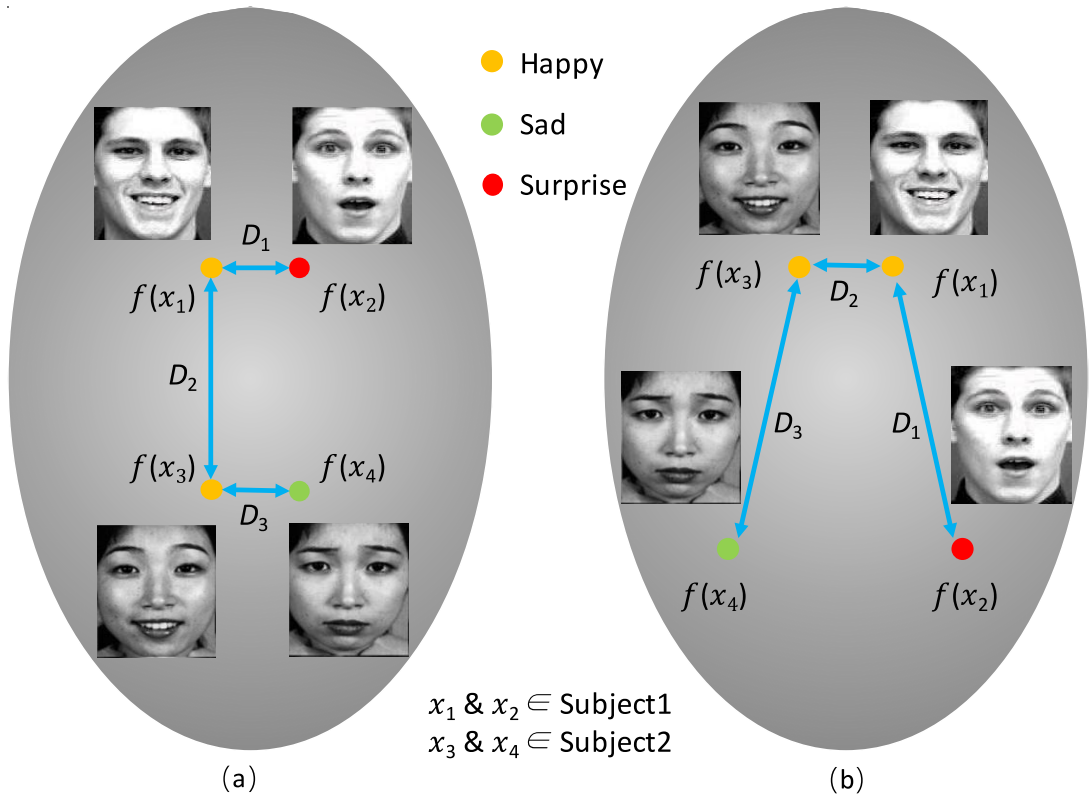

Figure 5.1. Illustration of representations in feature space learned by (a) existing methods, and (b) the proposed method.

Targeting for the problem of explicitly eliminating the detrimental variations following the prior knowledge of the task to achieve better generalization. It is challenging since the training set contains images annotated with multiple semantic variations of interest, but there is no example of the transformation ($e.g.,$ gender) as the unsupervised image translation [Dong, et al., 2017], [Li, et al., 2015], and the latent variation is totally unspecified [Liu, et al., 2019].

Following the terminology used in previous multi-class dataset (including a main-task label and several side-labels) [Jha, et al., 2018], [Makhzani, et al., 2015], [Mathieu, et al., 2016] , three complementary parts can be defined as in Fig.5.2.

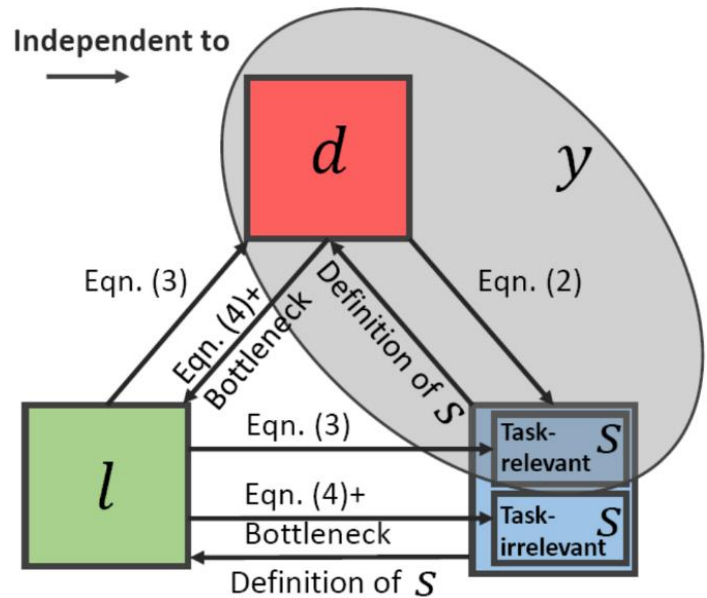

Figure 5.2. Illustration of the expected separations of the observation $x$, which associated with the discriminative representation d (red), latent variation l (green) and semantic variations ŝ (blue). Our framework explicitly enforces them marginally independent to each other. The $d$ and task-dependent $s$ are related to the main-recognition task label $y$.

The factors relate to the side-labels is named as the *semantic variations* ($s$), which can be either $task-relevant/irrelevant$ depending on whether they are marginally independent to the main recognition task or not. The *latent variation* ($l$) summarizes the remaining properties unspecified by main and semantic labels. How the DNN can systematically learn a *discriminative representation* ($d$) to be informative for the main recognition task, while marginally independent to multiple $s$ and unspecified $l$ in a controllable way remains challenging.

Several efforts have been made to enforce the main task representation invariant to a single task-*irrelevant* (independent) semantic factor, such as pose, expression or illumination-invariant face recognition via neural preprocessing [Huang, et al., 2017], [Tian, et al., 2018] or metric learning [Liu, et al., 2017].

To further improve the discriminating power of the expression feature representations, and address the large intra-subject variation in FER, a potential solution is to incorporate the deep metric learning scheme within a convolutional neural network (CNN) framework [Liu, et al., 2018], [Liu, et al., 2018], [Liu, et al., 2019], [Liu, et al., 2017]. The fundamental philosophy behind the widely-used triplet loss function [Ding, et al., 2015] is to require one positive example closer to the anchor example than one negative example with a fixed gap τ. Thus, during one iteration, the triplet loss ignores the negative examples from the rest of classes.

Moreover, one of the two examples from the same class in the triplets can be chosen as the anchor point. However, there exist some special cases that the triplet loss function with improper anchor may judge falsely, as illustrated in Fig.5.3. This means the performance is quite sensitive to the anchor selection in the triplets input. They adapted the idea from the (*N*+1)-tuplet loss [Sohn, 2016] and coupled clusters loss (CCL) [Liu, et al., 2016] to design a (*N*+*M*)-tuplet clusters loss function which incorporates a negative set with *N* examples and a positive set with *M* examples in a mini-batch. A reference distance *T* is introduced to force the negative examples to move away from the center of positive examples and for the positive examples to simultaneously map into a small cluster around their center $c^+$. The circles of radius $(T + \frac{\tau}{2})$ and $(T - \frac{\tau}{2})$ centered at the $c^+$ form the boundary of the negative set and positive set respectively, as shown in Fig.5.3.

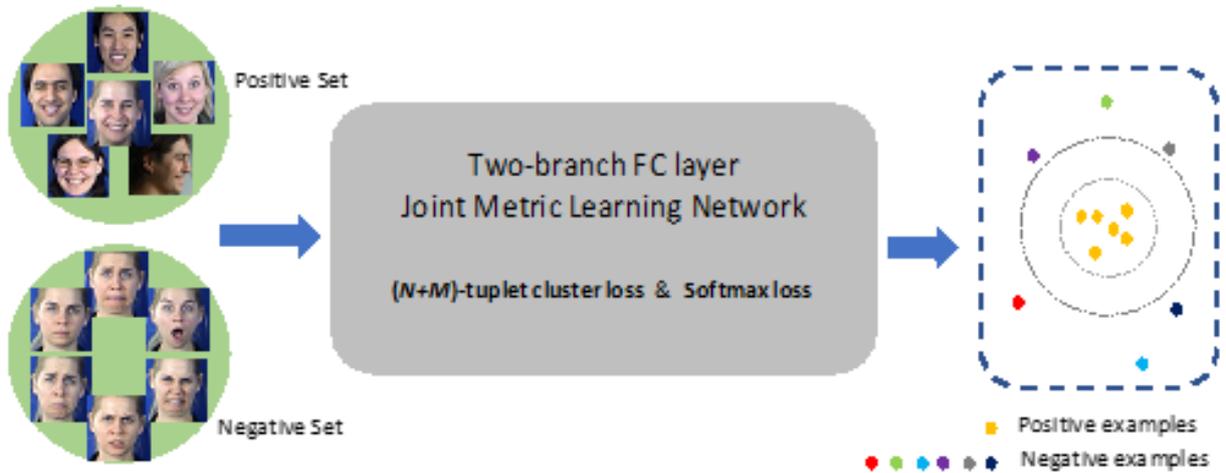

Figure 5.3. Framework of our facial expression recognition model used for training. The deep convolutional network aims to map the original expression images into a feature space that the images of the same expression tend to form a cluster while other images tend to locate far away

By doing this, it can handle complex distribution of intra- and inter-class variations, and free the anchor selection trouble in conventional deep metric learning methods. Furthermore, the reference distance *T* and the margin τ can be learned adaptively via the propagation in the CNN instead of the manually-set hyper-parameters. [Liu, et al., 2017] propose a simple and efficient mini-batch construction scheme that uses different expression images with the same identity as the negative set to avoid the expensive hard-negative example searching, while mining the positive set online.

Then, the (*N+M*)-tuplet clusters loss guarantees all the discriminating negative samples are efficiently used per update to achieve an identity-invariant FER. Besides, jointly optimize the softmax loss and (*N+M*)-tuplet clusters loss is used to explore the potential of both the expression labels and identity labels information. Considering the different characteristics of each loss function and their tasks, two branches of fully connected (FC) layers is developed, and a connecting layer to balance them. The features extracted by the expression classification branch can be fed to the following metric learning processing. This enables each branch to focus better on their own task without embedding much information of the other. As shown in Fig.5.3, the inputs are two facial expression image set: one positive set (images of the same expression from different subjects) and one negative set (images of other expressions with the same identity of the query example). The deep features and distance metrics are learned simultaneously in a network.

[Liu, et al., 2019], [Liu, et al., 2017] propose a generalized (*N+M*)-tuplet clusters loss function with adaptively learned reference threshold which can be seamlessly factorized into a linear-fully connected layer for an end-to-end learning. With the identity-aware negative mining and online positive mining scheme, the distance metrics can be learned with fewer input passes and distance calculations, without sacrificing the performance for identity-invariant FER. The softmax loss and (*N+M*)-tuplet clusters loss is optimized jointly in a unified two-branch FC layer metric learning CNN framework based on their characteristics and tasks. In experiments, [Liu, et al., 2017] demonstrate that the proposed method achieves promising results not only outperforming several state-of-art approaches in posed facial expression dataset (e.g., CK+, MMI), but also in spontaneous facial expression dataset (namely, SFEW).

However, the metric learning-based solution bears the drawback that the cost used to regularize the representation is pairwise [Liu], which does not scale well as the number of values that the attribute can take could be large [Liu, et al., 2019]. Since the invariance we care about can vary greatly across tasks, these approaches require us to design a new architecture each time when a new invariance is required.

Moreover, a basic assumption in their theoretical analysis is that the attribute is *irrelevant* to the prediction, which limits its capabilities in analyzing the task-*relevant* (dependent) semantic labels. These labels are usually used to achieve the attribute-enhanced recognition via the feature aggregation in multi-task learning [Hu, et al., 2017, Kingma and Ba, 2014], [Li, et al., 2018], [Peng, et al., 2017] (*e.g.*, gender, age and ethnicity can shrink the search space for face identification).

However, the invariance *w.r.t* those attributes are also desired in some specific tasks. For example, the makeup face recognition system should be invariant to age, hair color *etc*. Similarly, the gender and ethnicity are sensitive factors in fairness/bias-free classification when predicting the credit and health condition of a person. These semantic labels and the main task label are related due to the inherent bias within the data. A possible solution is setting this attribute as a random variable of a probabilistic model and reasoning about the invariance explicitly [Fu, et al., 2013, Liu, et al., 2015, Xiao, et al., 2017]. Since the divergence between a pair of distributions is used as the criteria to induce the invariance, the number of pairs to be processed grows quadratically with the number of attributes, which can be computationally expensive for the multiple variations in practice.

Another challenge is how to achieve better generalization by dispelling those latent variations without the label. For instance, we may expect the face recognition system not only be invariant to the expression following the side label, but also applicable to different race, which do not have side label. Noticing that this problem also share some similarity with feature disentanglements in image generation area [Guo, et al., 2013], [Makhzani, et al., 2015], while their goal is to improve content classification performance instead of synthesizing high-quality images.

Motivated by the aforementioned difficulties, [Liu, et al., 2019] proposes to enable a system which can dispel *a group of* undesired *task-irrelevant/relevant* and *latent* variations in an unsupervised manner: it does not need paired semantic transformation example [Dong, et al., 2017], [Li, et al., 2015] and latent labels.

Specifically, [Liu, et al., 2019] resort to an end-to-end conditional adversarial training framework. Their approach relies on an encoder-decoder architecture where, given an input image $x$ with its main-task label $y$ and to-be dispelled semantic variation label $s$, the encoders maps $x$ to a discriminative representation $d$ and a latent variation $l$, and the decoder is trained to reconstruct $x$ given ($d$,$s$,$l$). It configures a semantic discriminator condition to $s$ only, and two classifiers with inverse objectives which condition to $d$ and $l$, respectively to constrain the latent space for manipulating multiple variations for better scalability.

It is able to explicitly learn a task-specific discriminative representation with desired invariance property by systematically incorporating prior domain knowledge of the task. The to-be dispelled *multiple* semantic variations could be either task-*dependent/independent*, and the unspecified *latent* variation can also been eliminated in an *unsupervised* manner. Semantic discriminator and two inverse classifiers are introduced to constrain the latent space and result in a simpler training pipeline and better scalability. The theoretical equilibrium condition in different dependency scenarios have been analyzed. Extensive experiments on Extrended YaleB, 3 makeup set, CelebA, LFWA and DFW disgised face recognition benchmarks verifies its effectiveness and generality.

**5.2. Problem statement. Deep Metric learning based disentanglement for FER**

FER focus on the classification of seven basic facial expressions which are considered to be common among humans [Tian, et al., 2005]. Much progress has been made on extracting a set of features to represent the facial images [Jain, et al., 2011]. Geometric representations utilize the shape or relationship between facial landmarks. However, they are sensitive to the facial landmark misalignments [Shen, et al., 2015]. On the other hand, appearance features, such as Gabor filters, Scale Invariant Feature Transform (SIFT), Local Binary Patterns (LBP), Local Phase Quantization (LPQ), Histogram of Oriented Gradients (HOG) and the combination of these features via multiple kernel learning are usually used for representing facial textures [Baltrušaitis, et al., 2015], [Jiang, et al., 2011], [Yüce, et al., 2015], [Zhang, et al., 2014]. Some methods such as active appearance models (AAM) [Tzimiropoulos and Pantic, 2013] combine the geometric and appearance representations to provide better spatial information. Due to the limitations of handcrafted filters [Liu, et al., 2017], [Liu, et al., 2018], extracting purely expression-related features is difficult.

The developments in deep learning, especially the success of CNN [Liu], have made high-accuracy image classification possible in recent years [Che, et al., 2019], [Liu, et al., 2019], [Liu, et al., 2018], [Liu, et al., 2018]. It has also been shown that carefully designed neural network architectures perform well in FER [Mollahosseini, et al., 2016]. Despite its popularity, current softmax loss-based network does not explicitly encourage intra-class compactness and inter-class separation [Liu, et al., 2020], [Liu, et al., 2019]. The emerging deep metric learning methods have been investigated for person recognition and vehicle re-identification problems with large intra-class variations, which suggests that deep metric learning may offer more pertinent representations for FER [Liu, et al., 2019].

Compared to traditional distance metric learning, deep metric learning learns a nonlinear embedding of the data using the deep neural networks. The initial work is to train a Siamese network with contrastive loss function [Chopra, et al., 2005]. The pairwise examples are fed into two symmetric sub-networks to predict whether they are from the same class. Without the interactions of positive pairs and negative pairs, the Siamese network may fail to learn effective metrics in the presence of large intra- and inter-class variations. One improvement is the triplet loss approach [Ding, et al., 2015], which achieved promising performance in both re-identification and face recognition problems. The inputs are triplets, each consisting of a query, a positive example and a negative example. Specifically, it forces the difference of the distance from the anchor point to the positive example and from the anchor point to the negative example to be larger than a fixed margin $\tau$. Recently, some of its variations with faster and stable convergence have been developed. The most similar model of their proposed method is the ($N$+1)-tuplet loss [Sohn, 2016]. $x^+$ and $x^-$ denotong the positive and negative examples of a query example $x$, meaning that $x^+$ is the same class of $x$, while $x^-$ is not. Considering ($N$+1) tuplet which includes $x$, $x^+$ and $N$-1 negative examples $\{x_j^-\}_{j=1}^{N-1}$, the loss is:

$$L\left(x, x^+, \{x_j^-\}_{j=1}^{N-1}; f\right) = log\left(1 + \sum_{j=1}^{N-1} \exp(D(f, f^+) + \tau - D(f, f_j^-))\right) \quad (5.1)$$

where $f(\cdot)$ is an embedding kernel defined by the CNN, which takes $x$ and generates an embedding vector $f(x)$ and write it as $f$ for simplicity, with $f$ inheriting all superscripts and subscripts. $D(\cdot,\cdot)$ is defined as the Mahalanobis or Euclidean distance according to different implementations. The philosophy in this paper also shares commonality with the coupled clusters loss [Liu, et al., 2016], in which the positive example center $c^+$ is set as the anchor. By comparing each example with this center instead of each other mutually, the evaluation times in a mini-batch are largely reduced.

Despite their wide use, the above-mentioned frameworks still suffer from the expensive example mining to provide nontrivial pairs or triplets, and poor local optima [Liu, et al., 2019]. In practice, generating all possible pairs or triplets would result in quadratic and cubic complexity, respectively and the most of these pairs or triplets are less valuable in the training phase. Also, the online or offline traditional mini-batch sample selection is a large additional burden. Moreover, as shown in Fig.5.4 (a), (b) and (c), all of them are sensitive to the anchor point selection when the intra- and inter-class variations are large.

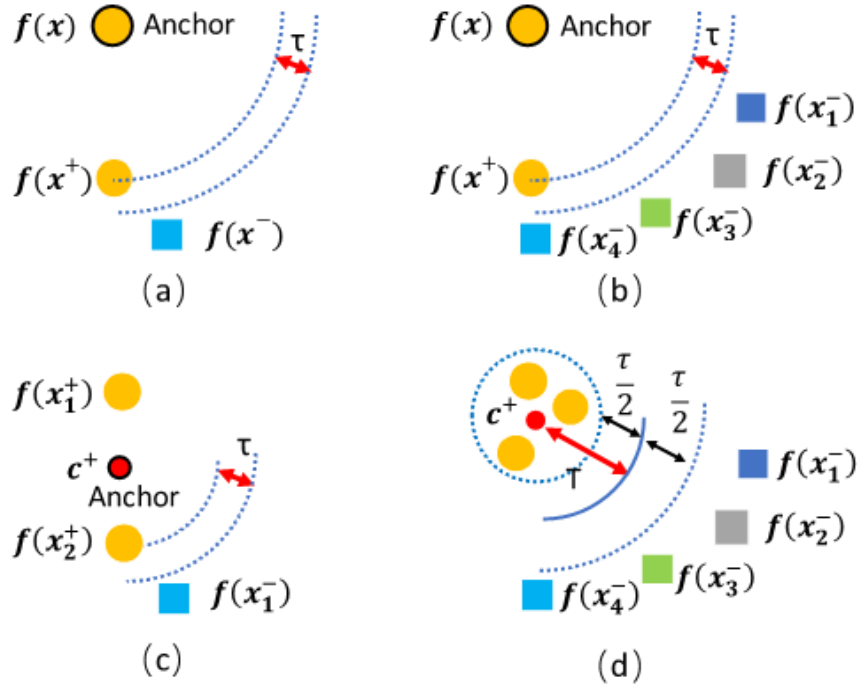

Figure 5.4. Failed case of (a) triplet loss, (b) (N+1)-tuplet loss, and (c) Coupled clusters loss. The proposed (N+M)-tuplet clusters loss is illustrated in (d).

The triplet loss, ($N$+1)-tuplet loss and CCL are 0, since the distances between the anchor and positive examples are indeed smaller than the distance between the anchor and negative examples for a margin τ. This means the loss function will neglect these cases during the back propagation. It need much more input passes with properly selected anchors to correct it. The fixed threshold in the contrastive loss was also proven to be sub-optimal for it failed to adapt to the local structure of data. Li et al. proposed [Li, et al., 2013] to address this issue by learning a linear SVM in a new feature space. Some works [Goodfellow, et al., 2013, Wang, et al., 2014] used shrinkage-expansion adaptive constraints for pair-wise input, which optimized by alternating between SVM training and projection on the cone of all positive semidefinite (PSD) matrices, but their mechanism cannot be implemented directly in deep learning.

A recent study presented objective comparisons between the softmax loss and deep metric learning loss and showed that they could be complementary to each other [Horiguchi, et al., 2016]. Therefore, an intuitive approach for improvement is combining the classification and similarity constraints to form a joint CNN learning framework. For example, [Sun, et al., 2014], [Yi, et al., 2014] combining the contrastive and softmax losses together to achieve a better performance, while [Zhang, et al., 2016] proposed to combine triplet and softmax loss via joint optimization. These models improve traditional CNN with softmax loss because similarity constraints might augment the information for training the network. The difficult learning objective can also effectively avoid overfitting. However, all these strategies apply the similarity as well as classification constraints directly on the last FC layer, so that harder tasks cannot be assigned to deeper layers, (i.e., more weights) and interactions between constraints are implicit and uncontrollable. Normally, the softmax loss converges much faster than the deep metric learning loss in multi-task networks. This situation has motivated us to construct a unified CNN framework to learn these two loss functions simultaneously in a more reasonable way.

### 5.3. Adversarial training based disentanglement

The task of Feature-level Frankenstein (FLF) framework can be formalized as follows: Given a training set $\mathcal{D} = \{x^1, s^1, y^1\}, \cdots, \{x^M, s^M, y^M\}$, of M samples $\{image, semantic\ variations, class\}$, we are interested in the task of disentangling the feature representation of x to be three complementary parts, i.e., discriminative representation d, semantic variation s and latent variation l. These three codes are expected to be marginally independent with each other, as illustrated schematically in Fig.5.2. In the case of face, typical semantic variations including gender, expressions $etc$. All the remaining variability unspecified by $y$ and $s$ fall into the latent part $l$. Note that there are two possible dependency scenarios of $s$ and $y$ as discussed in Sec. 1. This will not affect the definition of $l$, and the information related to $y$ should incorporate $d$ and some of the task-dependent $s$.

Multi-task learning is a typical method to utilize multi-class label. It has been observed in many prior works that jointly learning of main-task and *relevant* side tasks can help improve the performance in an aggregation manner [Hu, et al., 2017, Kingma and Ba, 2014, Li, et al., 2018, Peng, et al., 2017], while it are targeting for dispelling.

Generative Adversarial Networks (GANs) has aroused increasing attraction. Conventionally, under the two-player (*i.e.*, generator and discriminator) formulation, the vanilla GANs [Goodfellow, et al., 2014, Yang, et al., 2018] are good at generating realistic images, but their potential for recognition remains to be developed. The typical method use GANs as a preprocessing step of image, which is similar to the "denoise", and then use these processed images for normal training and testing [Huang, et al., 2017], [Liu, et al., 2017], [Lu, et al., 2017], [Netzer, et al., 2011], [Tenenbaum and Freeman, 2000], [Tian, et al., 2018], [Tzeng, et al., 2017]. [Liu, et al., 2019] deploy the trained network for predictions directly as a feature extractor.

Comparing with the pixel-level GANs [Huang, et al., 2017], [Lu, et al., 2017], [Tian, et al., 2018], [Xie, et al., 2017], their feature-level competition results in much simpler training schemes and nicely scales to multiple attributes. Moreover, they usually cannot dispel task-relevant $s$, $e.g.,$ dispel gender from identity cannot get verisimilar face image for subsequent network training [Yang, et al., 2019].

Besides, they usually focus on a single variation for a specific task. Actually, the most of GANs and adversarial domain adaptation [Cao, et al., 2018], [Li, et al., 2014], [Tishby and Zaslavsky, 2015] use binary adversarial objective and applied for no more than two distributions.

It is worth noting that some works of GANs, e.g., Semi-Supervised GAN [Kingma and Welling, 2013] and DR-GAN [Tian, et al., 2018] have claimed that they consider multiple side labels. Indeed, they have added a new branch for the multi-categorical classification, but their competing adversarial loss only confuses the discriminator by using two distributions (real or generated) and no adversarial strategies are adopted between different categories in the auxiliary multi-categorical classifier branch.

[Liu, et al., 2019] are different from them in two aspects: **1**) the input of semantic discriminator is feature, instead of real/synthesized image; **2**) the goal of encoder needs to match or align the feature distribution between any two different attributes, instead of only real/fake distribution, and there is no "real" class in semantic discriminator.

Fairness/bias-free classification also targets a representation that is invariant to certain task-relevant(dependent) factor (*i.e.*, bias) hence makes the predictions fair [Edwards and Storkey, 2015]. As data-driven models trained using historical data easily inherit the bias exhibited in the data, the Fair VAEs [Liu, et al., 2015] tackled the problem using a Variational Autoencoder structure [Kushwaha, et al., 2018] approached with maximum mean discrepancy (MMD) regularization [Li, et al., 2018]. [Xie, et al., 2017] proposed to regularize the $l_1$ distance between representation distributions of data with different nuisance variables to enforce fairness. These methods have the same drawback that the cost used to regularize the representation is pairwise, which does not scale well for multiple task-irrelevant semantic variations.

Latent variation disentangled representation is closely related to their work. It trying to separate the input into two complementary codes according to their correlation with the task for image transform in single label dataset setting [Bengio and Learning, 2009]. Early attempts [Simonyan and Zisserman, 2014] separate text from fonts using bilinear models. Manifold learning and VAEs were used in [Elgammal and Lee, 2004], [Kingma and Welling, 2013] to separate the digit from the style. "What-where" encoders [Zellinger, et al., 2017] combined the reconstruction criteria with discrimination to separate the factors that are relevant to the labels. Unfortunately, their approaches cannot be generalized to unseen identities. added the GAN's objective into the VAE's objective to relax this restriction using an intricate triplet training pipeline. [Bao, et al., 2018, Hadad, et al., 2018, Hu, et al., 2018, Jiang, et al., 2017, Liu, et al., 2018] further reduced the complexity. Inspired by them, [Liu, et al., 2019] make their framework implicitly invariant to unspecified $l$ for better generality in a simple yet efficient way, despite the core is dispel $s$ and do not target for image analogies [Makhzani, et al., 2015].

### 5.4. Methodology. Deep Metric learning based disentanglement for FER

Here give a simple description of the intuition to introduce a reference distance $T$ to control the relative boundary $(T-\frac{\tau}{2})$ and $(T+\frac{\tau}{2})$ for the positive and negative examples respectively, as shown in Fig.5.4 (d). The $(N+1)$-tuplet loss function in Eq.(5.1) can be rewritten as follows:

$$L\left(x, x^+, \{x_j^-\}_{j=1}^{N-1}; f\right) = log\left(1 + \sum_{j=1}^{N-1} exp(D(f, f^+) + (-T + \frac{\tau}{2} + T + \frac{\tau}{2}) - D(f, f_j^-))\right)$$
$$= log\left(1 + \sum_{j=1}^{N-1} exp(D(f, f^+) - T + \frac{\tau}{2}) * exp(T + \frac{\tau}{2} - D(f, f_j^-))\right) \quad (5.2)$$

Indeed, the $exp(D(f, f^+) - T + \frac{\tau}{2})$ term used to pull the positive example together and the $exp(T - \frac{\tau}{2} + D(f, f_j^-))$ term used to push the negative examples away have an "OR" relationship. The relatively large negative distance will make the loss function ignore the large absolute positive distance. One way to alleviate large intra-class variations is to construct an "AND" function for these two terms.

The triplet loss can also be extended to incorporate *N* negative examples and *M* negative examples. Considering a multi-classification problem, the triplet loss and CCL only compare the query example with one negative example, which only guarantees the embedding vector of the query one to be far from a selected negative class instead of every class. The expectation of these methods is that the final distance metrics will be balanced after sufficient number of iterations. However, towards the end of the training, individual iteration may exhibit zero errors due to the lack of discriminative negative examples causing the iterations to be unstable or slow in convergence.

The identity labels in FER database largely facilitate the hard-negative mining to alleviate the effect of the inter-subject variations. In practice, for a query example, [Liu, et al., 2017] compose its negative set with all the different expression images of the same person. Moreover, randomly choosing one or a group of positive examples is a paradigm of the conventional deep metric methods, but some extremely hard positive examples may distort the manifold and force the model to be over-fitting. In the case of spontaneous FER, the expression label may erroneously be assigned due to the subjectivity or varied expertise of the annotators [Barsoum, et al., 2016, Zafeiriou, et al., 2016]. Thus, an efficient online mining for *M* randomly-chosen positive examples should be designed for large intra-class variation datasets. [Liu, et al., 2017] find the nearest negative example and ignore those positive examples with a larger distance. Algorithm 1 shows the detail. In summary, the new loss function is expressed as follows:

$$L\left(\{x_i^+\}_{i=1}^M, \{x_j^-\}_{j=1}^N; f\right) = \frac{1}{M*}\sum_{i=1}^{M*} max(0, D(f^+, c^+) - T + \frac{\tau}{2})$$
$$+ \frac{1}{N}\sum_{j=1}^{N} max(0, T + \frac{\tau}{2} - D(f_j^-, c^+))) \quad (5.3)$$

The simplified geometric interpretation is illustrated in Fig.5.4 (d). Only if the distances from online mined positive examples to the updated $c^+$ smaller than $(T - \frac{\tau}{2})$ and the distances to the updated $c^+$ than $(T + \frac{\tau}{2})$, the loss can get a zero value. This is much more consistent with the principle used by many data cluster and discriminative analysis methods. One can see that the conventional triplet loss and its variations become the special cases of the (*N+M*)-tuplet clusters loss under their framework.

For a batch consisting of *X* queries, the input passes required to evaluate the necessary embedding feature vectors in the application are *X*, and the total number of distance calculations can be $2(N + M) * X$. Normally, the *N* and *M* are much smaller than *X*. In contrast, triplet loss requires $C_X^3$ passes and $2C_X^3$ times calculations, (*N+1*)- tuplet loss requires $(X + 1) * X$ passes and $(X + 1) * X^2$ times calculations. Even for a dataset with a moderate size, it is intractable to load all possible meaningful triplets into the limited memory for model training.

By assigning different values for $T$ and $\tau$, [Liu, et al., 2017] define a flexible learning task with adjustable difficulty for the network. However, the two hyper-parameters need manual tuning and validation. In the spirit of adaptive metric learning for SVM [Li, et al., 2013], [Liu, et al., 2017] formulate the reference distance to be a function $T(\cdot,\cdot)$ related with each example instead of a constant. Since the Mahalanobis distance matrix $\mathbf{M}$ in Eq.(5.4) itself is quadratic, and can be calculated automatically via a linear fully connected layer as in [Shi, et al., 2016], [Liu, et al., 2017] assume $T(f_1,f_2)$ as a simple quadratic form, i.e., $T(f_1,f_2)=\frac{1}{2}z^t\mathbf{Q}z + \omega^t z + b$, where $z^t = [f_1^t f_2^t] \in \mathbb{R}^{2d}$, $\mathbf{Q} = \begin{bmatrix} \mathbf{Q}_{f_1 f_1} & \mathbf{Q}_{f_1 f_2} \\ \mathbf{Q}_{f_2 f_1} & \mathbf{Q}_{f_2 f_2} \end{bmatrix} \in \mathbb{R}^{2d \times 2d}$, $\omega^t = [\omega_{f_1}^t \omega_{f_2}^t] \in \mathbb{R}^{2d}$, $b \in \mathbb{R}$, $f_1$ and $f_2 \in \mathbb{R}^{2d}$ are the representations of two images in the feature space.

$$D(f_1,f_2)=\|f_1 - f_2\|_M^2 = (f_1 - f_2)^T \mathbf{M}(f_1 - f_2) \tag{5.4}$$

Due to the symmetry property with respect to $f_1$ and $f_2$, $T(f_1,f_2)$ can be rewritten as follows:

$$T(f_1,f_2)=\tfrac{1}{2}f_1^t \widetilde{\mathbf{A}} f_1 + \tfrac{1}{2}f_2^t \widetilde{\mathbf{A}} n + f_1^t \widetilde{\mathbf{B}} f_2 + c^t(f_1 + f_2) + b \tag{5.5}$$

where $\widetilde{\mathbf{A}} = \mathbf{Q}_{f_1 f_1} = \mathbf{Q}_{f_2 f_2}$ and $\widetilde{\mathbf{B}} = \mathbf{Q}_{f_1 f_2} = \mathbf{Q}_{f_2 f_1}$ are both the $d \times d$ real symmetric matrices (not necessarily positive semi-definite), $c = \omega_{f_1} = \omega_{f_2}$ is a $d$-dimensional vector, and $b$ is the bias term. Then, a new quadratic formula $H(f_1,f_2)=T(f_1,f_2) - D(f_1,f_2)$ is defined to combine the reference distance function and distance metric function. Substituting Eq.(5.4) and Eq.(5.5) to $H(f_1,f_2)$, we get:

$$H(f_1,f_2)= \tfrac{1}{2}f_1^t(\widetilde{\mathbf{A}} - 2\mathbf{M})f_1 + \tfrac{1}{2}f_2^t(\widetilde{\mathbf{A}} - 2\mathbf{M})f + f_1^t(\widetilde{\mathbf{B}} + 2\mathbf{M})f_2 + c^t(f_1 + f_2) + b_2 \tag{5.6}$$

$$H(f_1,f_2)= \tfrac{1}{2}f_1^t \mathbf{A} f_1 + \tfrac{1}{2}f_2^t \mathbf{A} f_2 + f_1^t \mathbf{B} f_2 + c^t(f_1 + f_2) + b \tag{5.7}$$

where $\mathbf{A}=(\widetilde{\mathbf{A}} - 2\mathbf{M})$ and $\mathbf{B}=(\widetilde{\mathbf{B}} + 2\mathbf{M})$. Suppose $\mathbf{A}$ is positive semi-definite (PSD) and $\mathbf{B}$ is negative semi-definite (NSD), $\mathbf{A}$ and $\mathbf{B}$ can be factorized as $\mathbf{L}_A^T \mathbf{L}_A$ and $\mathbf{L}_B^T \mathbf{L}_B$. Then $H(f_1,f_2)$ can be formulated as follows:

$$H(f_1,f_2)= \tfrac{1}{2}f_1^t \mathbf{L}_A^T \mathbf{L}_A f_1 + \tfrac{1}{2}n^t \mathbf{L}_A^T \mathbf{L}_A f_2 + f_1^t \mathbf{L}_B^T \mathbf{L}_B f_2 + c^t(f_1 + f_2) + b$$
$$= \tfrac{1}{2}(\mathbf{L}_A f_1)^t (\mathbf{L}_A f_1) + \tfrac{1}{2}(\mathbf{L}_A f_2)^t (\mathbf{L}_A f_2) + (\mathbf{L}_B f_1)^t (\mathbf{L}_B f_2) + c^t f_1 + c^t f_2 + b \tag{5.8}$$

Motivated by the above, [Liu, et al., 2017] propose a general, computational feasible loss function. Following the notations in the preliminaries and denote $(\mathbf{L}_A, \mathbf{L}_B, c)^T$ as $W$:

$$L\left(W, \{x_i^+\}_{i=1}^M, \{x_j^-\}_{j=1}^N; f\right) =$$

$$\tfrac{1}{M*}\Sigma_{i=1}^{M*}\max(0, H(f_i^+, c^+)+\tfrac{\tau}{2}) + \tfrac{1}{N}\Sigma_{j=1}^{N}\max(0, H(f_j^-, c^+) + \tfrac{\tau}{2} \quad (5.9)$$

Given the mined $N+M*$ training examples in a mini-batch, $l(\cdot)$ is a label function. If the example $x_k$ is from the positive set, $l(x_k) = -1$, otherwise, $l(x_k) = 1$. Moreover, the $\tfrac{\tau}{2}$ can be simplified to be the constant 1, and changing it to any other positive value results only in the matrices being multiplied by corresponding factors. The hinge-loss like function is:

$$L\left(W, \{x_i^+\}_{i=1}^{M}, \{x_j^-\}_{j=1}^{N}; f\right) = \tfrac{1}{N+M*}\Sigma_{k=1}^{N+M*}\max(0, l(x_k)*H(f_k, c^+)+1) \quad (5.10)$$

[Liu, et al., 2017] optimize Eq.(5.10) using the standard stochastic gradient descent with momentum. The desired partial derivatives of each example are computed as:

$$\frac{\partial L}{\partial W^l} = \frac{1}{N+M*}\Sigma_{k=1}^{N+M*} \frac{\partial L}{\partial X_k^l} \frac{\partial X_k^l}{\partial W^l} \quad (5.11)$$

$$\frac{\partial L}{\partial X_k^l} = \frac{\partial L}{\partial X_k^{l+1}} \frac{\partial X_k^{l+1}}{\partial X_k^l} \quad (5.12)$$

where $X_k^l$ represents the feature map of the example $x_k$ at the $l_{th}$ layer. Eq.(5.11) shows that the overall gradient is the sum of the example-based gradients. Eq.(5.12) shows that the partial derivative of each example with respect to the feature maps can be calculated recursively. So, the gradients of network parameters can be obtained with back propagation algorithm.

In fact, as a straightforward generalization of conventional deep metric learning methods, the ($N+M$)-tuplet clusters loss can be easily used as a drop-in replacement for the triplet loss and its variations, as well as used in tandem with other performance-boosting approaches and modules, including modified network architectures, pooling functions, data augmentations or activation functions.

The proposed two-branch FC layer joint metric learning architecture with softmax loss and ($N+M$)-tuplet clusters loss, denoted as 2B($N+M$)Softmax. The convolutional groups of the network are based on the inception FER network presented in [Mollahosseini, et al., 2016]. [Liu, et al., 2017] adopt the parametric rectified linear unit (PReLU) to replace the conventional ReLU for its good performance and generalization ability when given limited training data. In addition to providing the sparsity to gain benefits discussed in [Arora, et al., 2014], the inception layer also allows for improved recognition of local features. The locally applied smaller convolution filters seem to align the way that human process emotions with the deformation of local muscles.

Combing the ($N+M$)-tuplet clusters loss and softmax loss is an intuitive improvement to reach a better performance. However, conducting them directly on the last FC layer is sub-optimal. The basic idea of building a two-branch FC layers after the deep convolution groups is combining two losses in different level of tasks. [Liu, et al., 2017] learn the detailed features shared between the same expression class with the expression classification (EC) branch, while exploiting semantic representations via the metric learning (ML) branch to handle the significant appearance changes from different subjects. The connecting layer embeds the information learned from the expression label-based detail task to the identity label-based semantical task, and balances the scale of weights in two task streams. This type of combination can effectively alleviate the interference of identity-specific attributes. The inputs of connecting layer are the output vectors of the former FC layers- $FC_2$ and $FC_3$, which have the same dimension denoted as $D_{\text{input}}$. The output of the connecting layer, denoted as $FC_4$ with dimension $D_{\text{ouput}}$, is the feature vector fed into the second layer of the ML branch. The connecting layer concatenates two input feature vectors into a larger vector and maps it into a $D_{\text{output}}$ dimension space:

$$FC_4 = \mathbf{P}^{\text{T}}[FC_2 ; FC_3] = \mathbf{P}_1^{\text{T}} FC_2 + \mathbf{P}_2^{\text{T}} FC_3 \tag{5.13}$$

where $\mathbf{P}$ is a $2(D_{\text{input}} \times D_{\text{output}})$ matrix, $\mathbf{P_1}$ and $\mathbf{P_2}$ are $D_{\text{input}} \times D_{\text{output}}$ matrices.

Regarding the sampling strategy, every training image is used as a query example in an epoch. In practice, the softmax loss will only be calculated for the query example. The importance of two loss functions is balanced by a weight α. During the testing stage, this framework takes one facial image as input, and generates the classification result through the EC branch with the softmax loss function.

**5.5. Adversarial training based disentanglement**
**5.5.1. The structure of representations**

For the latent variation encoding, [Liu, et al., 2019] choose the $l$ to be a vector of real value rather than a one-hot or a class ordinal vector to enable the network to be generalized to identities that are not presented in the training dataset as in [Bao, et al., 2018, Makhzani, et al., 2015]. However, as the semantic variations are human-named for a specific domain, this concern is removed. In theory, $s$ can be any type of data (*e.g.*, continuous value scalar/vector, or a sub-structure of a natural language sentence) as long as it represents a semantic attribute of *x* under their framework. For simplicity, [Liu, et al., 2019] consider here the case where $s$ is a *N*-dimensional binary variable for *N* to-be controlled semantic variations. Regarding the multi-categorical labels, they are factorized to multiple binary choices. The domain adaptation could be a special case of their model when the semantic variation is the Bernoulli variable which takes the one-dimensional binary value (*i.e.*, $s = \{0,1\}$), representing the domains.

**5.5.2. Framework architecture**

The model described in Fig.5.5 is proposed to achieve the objective based on an encoder-decoder architecture with conditional adversarial training. At inference time, a test image is encoded to the $d$ and $l$ in the latent space, and the $d$ can be used for recognition task with desired invariant property $w.r.t.$ the $s$. Besides, the user can choose the combination of $(d,s,l)$ that are fed to the decoder for different image transforms.

156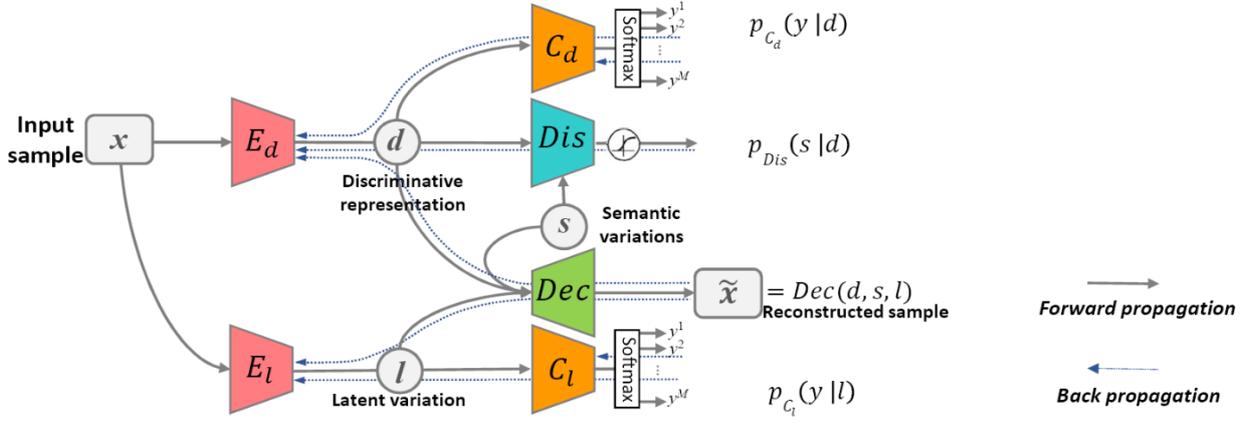

Figure 5.5. Failed case of (a) triplet loss, (b) (N+1)-tuplet loss, and (c) Coupled clusters loss. The proposed (N+M)-tuplet clusters loss is illustrated in (d).

**5.5.3. Informative to main-recognition task**. The discriminative encoder $E_d$ with parameter $\theta_{E_d}$ maps an input image to its discriminative representation $d = E_d(x)$ which is informative for the main recognition task and invariant to some semantic attributes. By invariance, we mean that given two samples $x^1$, $x^2$ from a subject class ($y^1 = y^2$) but with different semantic attribute labels ($s^1 \neq s^2$), their $d^1$ and $d^2$ are expected to be the same. Given the obtained $d$, [Liu, et al., 2019] expect to predict its corresponding label $y$ with the classifier $C_d$ to model the distribution $p_{C_d}(y|x)$. The task of $C_d$ and the first objective of the $E_d$ is to ensure the accuracy of the main recognition task. Therefore, [Liu, et al., 2019] update them to minimize:

$$\min_{E_d, C_d} \mathcal{L}_{C_d} = \mathbb{E}_{x,y \sim q(x,s,y)} \text{-log} p_{C_d}(y|E_d(x)) \quad (5.14)$$

where the categorical cross-entropy loss is used for the classifier. The $q(x, s, y)$ is the true underlying distribution that the empirical observations are drawn from.

**5.5.4. Eliminating semantic variations**. The discriminator $Dis$ output probabilities of an attribute vector $p_{Dis}(s|d)$. In practical implementation, this is made by concatenating $d$ and binary attributes code $s$ for input and outputs the [0,1] values using the sigmoid unit. Its loss depends on the current state of semantic encoders and is written as:

$$\min_{Dis} \max_{E_d} \mathcal{L}_{Dis} = \mathbb{E}_{x,s \sim q(x,s,y)} \text{-log} p_{Dis}(s|E_d(x)) \quad (5.15)$$

Concretely, the $Dis$ and $E_d$ form an adversarial game, in which the $Dis$ is trained to detect an attribute of data by maximizing the likelihood $p_{Dis}(s|d)$, while the $E_d$ fights to conceal it by minimizing the same likelihood. Eq. 15 guarantees that $d$ is marginally independent to $s$. Supposing that a semantic variation follows the Bernoulli distribution, the loss is formulated as $-\{s \log Dis(d) + (1-s)\log(1 - Dis(d))\}$. The proposed framework is readily amenable to control multiple attributes by extending the dimension of semantic variation vector. With *N* to-be dispelled semantic variations, $\log p_{Dis}(s|d) = \sum_{i=1}^{N} \{\log p_{Dis}(s_i|d)\}$. Note that even with binary attribute values at the training stage, each attribute can be considered as a continuous variable during inference to choose how much a specific attribute is perceivable in the generated images.

As discussed above, the semantic discriminator is essentially different from conventional GANs. The feature-level competition also similar to adversarial auto-encoder [Maaten and Hinton, 2008], which match the intermediate feature with a prior distribution (Gaussian). However, it is conditioned to another vector *s*, and require the encoder align the distribution between any two *s*, instead of only real/fake.

**5.5.5. Eliminating latent variation**. To train the latent variation encoder $E_l$, [Liu, et al., 2019] propose a novel variant of adversarial networks, in which the $E_l$ plays a minimax game with a classifier $C_l$ instead of a discriminator. The $C_l$ inspects the background latent variation *l* and learns to predict class label correctly, while the $E_l$ is trying to eliminate task-specific factors *d* by fooling $C_l$ to make false predictions.

$$\min_{C_l} \max_{E_l} \mathcal{L}_{C_l} = \mathbb{E}_{x,y \sim q(x,s,y)} \text{-}\log p_{C_l}(y|E_l(x)) \quad (5.16)$$

Since the ground truth of *d* is unobservable, [Liu, et al., 2019] use the *y* in here, which incorporate *d* and main-task relevant *s*. [Liu, et al., 2019] also use softmax output unit and cross-entropy loss in their implementations. In contrast to using three parallel VAEs [Makhzani, et al., 2015], the adversarial classifiers are expected to alleviate the costly training pipeline and facilitate the convergence.

**5.5.6. Complementary constraint**. The decoder *Dec* is a deconvolution network to produce a new version of the input image given the concatenated codes $(d, s, l)$. These three parts should contain enough information to allow the reconstruction of the input *x*. Herein, [Liu, et al., 2019] measure the similarity of the reconstruction with the self-regularized mean squared error (MSE) for simply:

$$\min_{E_d, E_l, Dec} \mathcal{L}_{rec} = \mathbb{E}_{x,s,y \sim q(x,s,y)} \|Dec(d, s, l) - x\|_2^2 \quad (5.17)$$

This design contributes to variation separation in an implicit way, and makes the encoded features more inclusive of the image content.

**5.6. Experiments and analysis.**
**5.6.1. Deep Metric learning based disentanglement for FER**

For a raw image in the database, face registration is a crucial step for good performance. The bidirectional warping of Active Appearance Model (AAM) [30] and a Supervised Descent Method (SDM) called IntraFace model [45] are used to locate the 49 facial landmarks. Then, face alignment is done to reduce in-plane rotation and crop the region of interest based on the coordinates of these landmarks to a size of 60 × 60. The limited images of FER datasets is a bottleneck of deep model implementation. Thus, an augmentation procedure is employed to increase the volume

158158158158

|  | | Predict | | | | | | |
|---|---|---|---|---|---|---|---|---|
|  | | AN | CO | DI | FE | HA | SA | SU |
| Actual | AN | **91.1%** | 0% | 0% | 1.1% | 0% | 7.8% | 0% |
|  | CO | 5.6% | **90.3%** | 0% | 2.7% | 0% | 5.6% | 0% |
|  | DI | 0% | 0% | **100%** | 0% | 0% | 0% | 0% |
|  | FE | 0% | 4% | 0% | **98%** | 2% | 0% | 8% |
|  | HA | 0% | 0% | 0% | 0% | **100%** | 0% | 0% |
|  | SA | 3.6 | 0% | 0% | 1.8% | 0% | **94.6%** | 0% |
|  | SU | 0% | 1.2% | 0% | 0% | 0% | 0% | **98.8%** |

Table 5.1. Average confusion matrix obtained from proposed method on the CK+ database.

of training data and alleviate the chance of over-fitting. [Liu, et al., 2017] randomly crop the 48×48 size patches, flip them horizontally and transfer them to grayscale images. All the images are processed with the standard histogram equalization and linear plane fitting to remove unbalanced illumination. Finally, [Liu, et al., 2017] normalize them to a zero mean and unit variance vector. In the testing phase, a single center crop with the size of 48×48 is used as input data.

Following the experimental protocol in [Mollahosseini, et al., 2016], [Yu and Zhang, 2015], [Liu, et al., 2017] pre-train their convolutional groups and EC branch FC layers on the FER2013 database [Goodfellow, et al., 2013] for 300 epochs, optimizing the softmax loss using stochastic gradient decent with a momentum of 0.9. The initial network learning rate, batch size, and weight decay parameter are set to 0.1, 128, 0.0001, respectively. If the training loss increased more than 25% or the validation accuracy does not improve for ten epochs, the learning rate is halved and the previous network with the best loss is reloaded. Then the ML branch is added and the whole network is trained by 204,156 frontal viewpoints (-45° to 45°) face images selected from the CMU Multi-pie [Gross, et al., 2010] dataset. There contains 337 people displaying disgust, happy, surprise and neutron. The size of both the positive and negative set are fixed to 3 images. The weights of two loss functions are set equally. [Liu, et al., 2017] select the highest accuracy training epoch as the pre-trained model.

In the fine-tuning stage, the positive and negative set size are fixed to 6 images (for CK+ and SFEW) or 5 images (for MMI). For a query example, the random searching is employed to select the other 6 (or 5) same expression images to form the positive set. Identity labels are required for negative mining in their method. CK+ and MMI have the subject IDs while the SFEW need manually label. In practice, an off-the-shelf face recognition method can be used to produce this information. When the query example lacks some expression images from the same subject, the corresponding expression images sharing the same ID with the any other positive examples are used. The tuplet-size is set to 12, which means 12×(6+6) =144 (or 12×(5+5) =120) images are fed in each training iteration. [Liu, et al., 2017] use Adam [Kingma and Ba, 2014] for stochastic optimization and other hyper-parameters such as learning rate are tuned accordingly via cross-validation. All the CNN architectures are implemented with the widely used deep learning tool "Caffe [Jia, et al., 2014]."

|  |  | \multicolumn{6}{c}{Predict} |
|  |  | AN | DI | FE | HA | SA | SU |
| --- | --- | --- | --- | --- | --- | --- | --- |
| Actual | AN | **81.8%** | 3% | 3% | 1.5% | 10.6% | 0% |
|  | DI | 10.9% | **71.9%** | 3.1% | 4.7% | 9.4% | 6% |
|  | FE | 5.4% | 8.9% | **41.4%** | 7.1% | 7.1% | 30.4% |
|  | HA | 1.1% | 3.6% | 0% | **92.9%** | 2.4% | 0% |
|  | SA | 17.2% | 7.8% | 0% | 1.6% | **73.4%** | 0% |
|  | SU | 7.3% | 0% | 14.6% | 0% | 0% | **79.6%** |

Table 5.2. Average confusion matrix obtained from proposed method on the MMI database.

To evaluate the effectiveness of the proposed method, extensive experiments have been conducted on three well-known publicly available facial expression databases: CK+, MMI and SFEW. For the fair comparison, [Liu, et al., 2017] follow the protocol used by previous works [Mollahosseini, et al., 2016, Yu and Zhang, 2015]. Three baseline methods are employed to demonstrate the superiority of the novel metric learning loss and two-branch FC layer network respectively, i.e., adding the ($N+M$)-tuplet clusters loss or ($N+1$)-tuplet loss with softmax loss after the EC branch, denoted as 1B($N+1$)Softmax or 1B($N+M$)Softmax, and combining the ($N+1$)-tuplet loss with softmax loss via the two-branch FC layer structure, as 2B($N+1$)Softmax. With randomly selected triplets, the loss failed to converge during the training phase.

The extended Cohn-Kanade database (CK+) [Lucey, et al., 2010] includes 327 sequences collected from 118 subjects, ranging from 7 different expressions (i.e., anger, contempt, disgust, fear, happiness, sadness, and surprise). The label is only provided for the last frame (peak frame) of each sequence. [Liu, et al., 2017] select and label the last three images, and obtain 921 images (without neutral). The final sequence-level predictions are made by selecting the class with the highest possibility of the three images. [Liu, et al., 2017] split the CK+ database to 8 subsets in a strict subject independent manner, and an 8-fold cross-validation is employed. Data from 6 subsets is used for training and the others are used for validation and testing. The confusions matrix of the proposed method evaluated on the CK+ dataset is reported in Table 1. It can be observed that the disgust and happy expressions are perfectly recognized while the contempt expression is relatively harder for the network because of the limited training examples and subtle muscular movements. As shown in Table 3, the proposed 2B($N+M$)Softmax outperforms the human-crafted feature-based methods, sparse coding-based methods and the other deep learning methods in comparison. Among them, the 3DCNN-DAP, STM-Explet and DTAGN utilized temporal information extracted from sequences. Not surprisingly, it also beats the baseline methods obviously benefit from the combination of novel deep metric learning loss and two-branch architecture.

The MMI database [Pantic, et al., 2005] includes 31 subjects with frontal-view faces among 213 image sequences which contain a full temporal pattern of expressions, i.e., from neutral to one of six basic expressions as time goes on, and then released. It is especially favored by the video-based methods to exploit temporal information. [Liu, et al., 2017] collect three frames in the middle of each image sequence and associate them with the labels, which results in 624 images in their experiments. [Liu, et al., 2017] divide MMI dataset into 10 subsets for person-independent ten-fold cross validation. The sequence-level predictions are obtained by choosing the class with the highest average score of the three images. The confusion matrix of the proposed method on the MMI database is reported in Table 2. As shown in Table 3, the performance improvements in this small database without causing overfitting are impressive. The proposed method outperforms other works that also use static image-based features and can achieve comparable and even better results than those video-based approaches.

Table 5.3.

Recognition accuracy comparison on the CK+ database [26] in terms of seven expressions, MMI database [Pantic, et al., 2005] in terms of six expressions, and SFEW database in terms of seven expressions.

| Methods | CK+ | MMI | Methods | SFEW |
|---|---|---|---|---|
| MSR [33] | 91.4% | N/A | Kim et al. [Kim, et al., 2016] | 53.9% |
| ITBN [44] | 91.44% | 59.7% | Ng et al. [Ng, et al., 2015] | 48.5% |
| BNBN [25] | 96.7% | N/A | Yao et al. [Yao, et al., 2015] | 43.58% |
| IB-CNN [11] | 95.1% | N/A | Sun et al. [Sun, et al., 2015] | 51.02% |
| 3DCNN-DAP [23] | 92.4% | 63.4% | Zong et al. [Zong, et al., 2015] | N/A |
| STM-Explet [24] | 94.19% | 75.12% | Kaya et al. [Kaya and Salah, 2016] | 53.06% |
| DTAGN [16] | 97.25% | 70.2% | Mao et al. [Mao, et al., 2016] | 44.7% |
| Inception [28] | 93.2% | 77.6% | Mollahosseini [Mollahosseini, et al., 2016] | 47.7% |
| 1B($N$+1)Softmax | 93.21% | 77.72% | 1B($N$+1)Softmax | 49.77% |
| 2B($N$+1)Softmax | 94.3% | 78.04% | 2B($N$+1)Softmax | 50.75% |
| 1B($N$+M)Softmax | 96.55% | 77.88% | 1B($N$+$M$)Softmax | 53.36% |
| **2B($N$+$M$)Softmax** | **97.1%** | **78.53%** | **2B($N$+$M$)Softmax** | **54.19%** |

Table 5.4. Average confusion matrix obtained from proposed method on the SFEW validation set.

|  |  | Predict | | | | | | |
|---|---|---|---|---|---|---|---|---|
|  |  | AN | DI | FE | HA | NE | SA | SU |
| Actual | AN | **66.24%** | 1.3% | 0% | 6.94% | 9.09% | 5.19% | 10.69% |
|  | DI | 21.74% | **4.35%** | 4.35% | 30.34% | 13.04% | 4.35% | 21.74% |
|  | FE | 27.66% | 0% | **6.38%** | 8.51% | 10.64% | 19.15% | 27.66% |
|  | HA | 0% | 0% | 0% | **87.67%** | 6.85% | 1.37% | 4.11% |
|  | NE | 5.48% | 0% | 2.74% | 1.37% | **57.53%** | 5.48% | 27.4% |
|  | SA | 22.81% | 0% | 1.75% | 7.02% | 8.77% | **40.35%** | 19.3% |
|  | SU | 1.16% | 0% | 2.33% | 5.81% | 17.44% | 0% | **73.26%** |

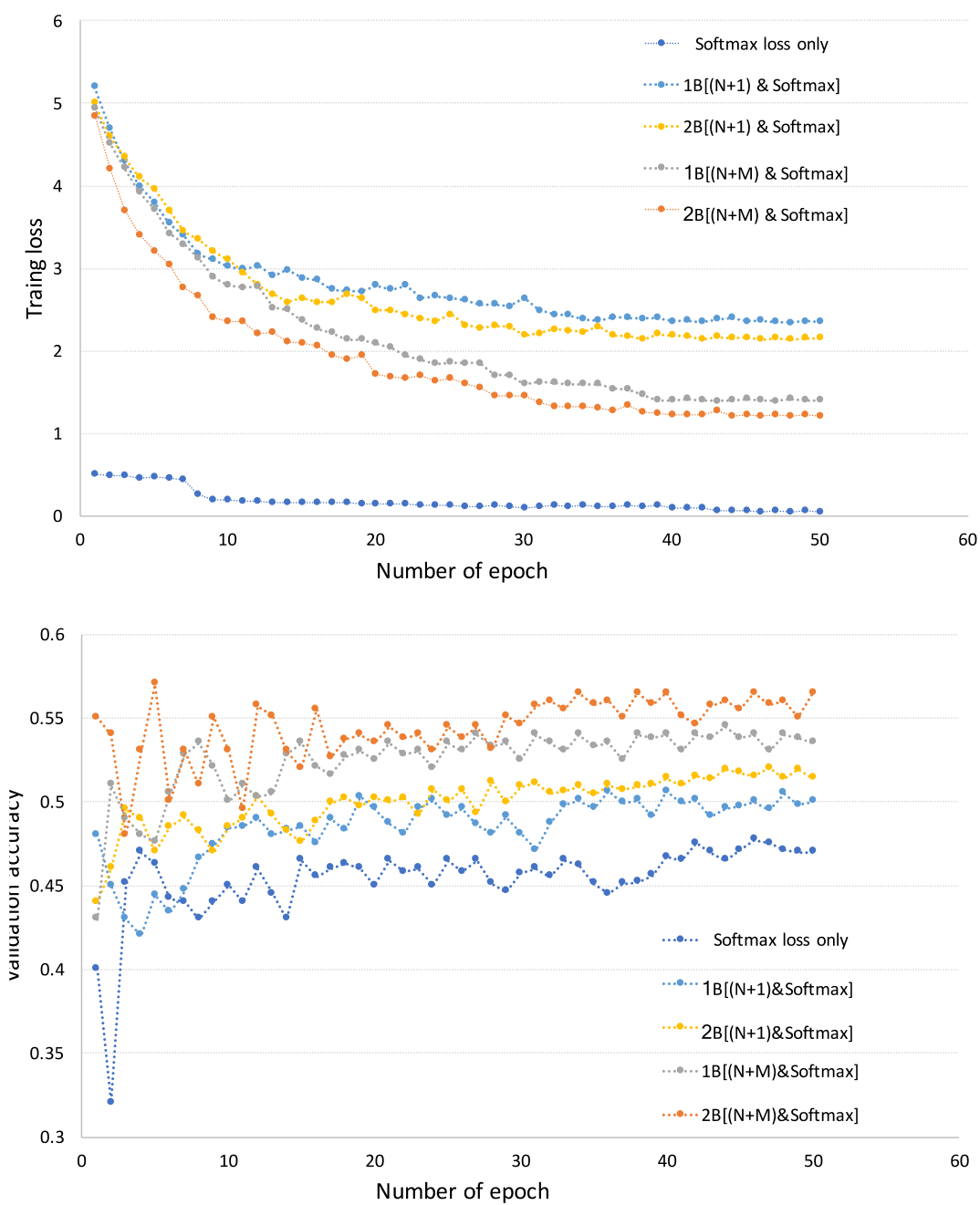

Figure 5.6. The training loss of different methods on SFEW validation set. The validation accuracies of different methods on SFEW validation set.

The static facial expressions in the wild (SFEW) database [Dhall, et al., 2015] is created by extracting frames from the film clips in the AFEW data corpus. There are 1766 well-labeled images (i.e., 958 for training, 436 for validation and 372 for testing) being assigned to be one of the 7 expressions. Different from the previous two datasets, it targets for unconstrained facial expressions, which has large variations reflecting real-world conditions. The confusion matrix of their method on the SFEW validation set is reported in Table 5.3. The recognition accuracy of disgust and fear are much lower than the others, which is also observed in other works. As illustrated in Table 4, the CNN-based methods dominate the ranking list. With the augmentation of deep metric learning and two-branch FC layer network, the proposed method works well in the real world environment setting. Note that Kim et al. [Kim, et al., 2016] employed 216 AlexNet-like CNNs with different architectures to boost the final performance. Their network performs about 25M operations, almost four times fewer than a single AlexNet. With the smaller size, the evaluation time in testing phase takes only 5ms using a Titan X GPU, which makes it applicable for real-time applications.

Overall, one can see that joint optimizing the metric learning loss and softmax loss can successfully capture more discriminative expression-related features and translate them into the significant improvement of FER accuracy. The ($N+M$)-tuplet clusters loss not only inherits merits of conventional deep metric learning methods, but also learns features in a more efficient and stable way. The two-branch FC layer can further give a boost in performance. Some nice properties of the proposed method are verified by Fig.5.6, where the training loss of 2B($N+M$)Softmax converges after about 40 epochs with a more steady decline and reaches a lower value than those baseline methods as expect. As Fig.5.7 illustrates, the proposed method and the baseline methods achieve better performance in terms of the validation accuracy on the training phase.

### 5.6.2. Adversarial training-based disentanglement

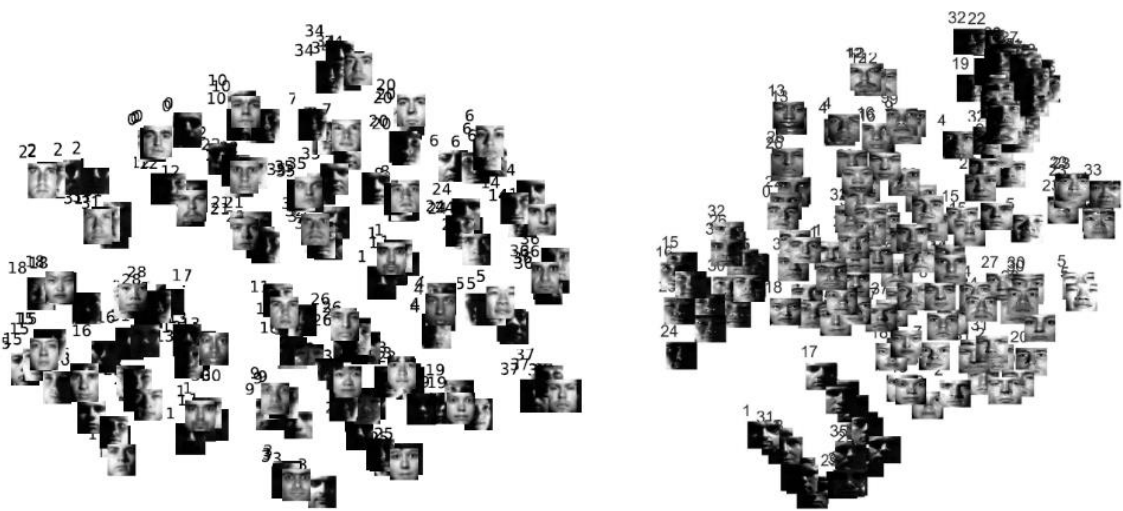

Figure 5.7. t-SNE visualization of images in Extended YaleB. The original images (b) are clustered according to their lighting environments, while the discriminative representation learned by our framework (a) is more likely to cluster with only identities.

To illustrate the behavior of the **F**eature-**l**evel **F**rankenstein (FLF) framework, [Liu, et al., 2019]quantitatively evaluate the discriminative representation with desired invariance property on three different recognition tasks and also offer qualitative evaluations by visually examining the perceptual quality of conditional face generation. As the frequent metrics (*e.g.*, log-likelihood of a set of validation samples) are not meaningful for perceptual generative models [Sun, et al., 2017], [Liu, et al., 2019] measure the information associated with the semantic variations $s$ or main-task label $y$ that is contained in each representation part to evaluate the degree of disentanglement as in [Liu, et al., 2015, Makhzani, et al., 2015].

In all experiments, [Liu, et al., 2019] utilize the Adam optimization method [Kingma, et al., 2014] with a learning rate of 0.001 and beta of 0.9 for the training of the encoders-decoder network, discriminator and classifiers. [Liu, et al., 2019] use a variable weight for the discriminator loss coefficient $\alpha$. [Liu, et al., 2019] initially set $\alpha$ to 0 and the model is trained as a normal auto-encoder. Then, $\alpha$ is linearly increased to 0.5 over the first 500,000 iterations to slowly encourage the model to produce invariant representations. This scheduling turned out to be critical in their experiments. Without it, they observed that the $E_d$ was too affected by the loss coming from the discriminator, even for low values of $\alpha$. All the models were implemented using TensorFlow.

For our lighting-tolerant classification task, [Liu, et al., 2019] use the Extended Yale B dataset [Georghiades, et al., 2001]. It comprises face images from 38 subjects under 5 different lighting conditions, *i.e.*, front, upper left, upper right, lower left, or lower right. [Liu, et al., 2019]aim to predict the subject identity $y$ using $d$. The semantic variable $s$ to be purged here is the lighting condition, while the latent variation $l$ does not have practical meaning in this dataset setting. [Liu, et al., 2019] follow the two-layer $E_d$ structure and train/test split of [Li, et al., 2018, Liu, et al., 2015]. 190 samples are utilized for training and all remaining 1,096 images are used for testing.

The numerical results of recognition using $E_d$ and $C_d$ are shown in Table 5. [Liu, et al., 2019] compare it with the state-of-the-art methods that use MMD regularizations *etc.,* to remove the affects of lighting conditions [Li, et al., 2018, Liu, et al., 2015]. The advantage of their framework about factoring out lighting conditions is shown by the improved accuracy 90.1%, while the best baseline achieves an accuracy of 86.6%. Although the lighting conditions can be modeled very well with a Lambertian model, [Liu, et al., 2019] choose to use a generic neural network to learn invariant features, so that the proposed method can be readily applied to other applications.

In terms of removing $s$, their framework can filter the lighting conditions since the accuracy of classifying $s$ from $d$ drops from 56.5% to 26.2% (halved), as shown in Table 5. Note that 20% is a chance performance for 5 class illumination, when the $s$ is totally dispelled. This can also be seen in the visualization of two-dimensional embeddings of the original $x$. One can see that the original images are clustered based on the lighting conditions. The clustering based on CNN features are almost well according to the identity, but still affected by the lighting and results in a 'black center'. As soon as removing the lighting variations via FLF, images are distributed almost only according to the identity of each subject.

Table. 5 Classification accuracy comparisons. Expecting the accuracy of classifying $y$ or $s$ from $l$ to be a lower value. A better discriminative representation $d$ has a higher accuracy of classifying $y$ and a lower accuracy in predicting $s$. *Following the setting in [Liu, et al., 2015], [Liu, et al., 2019] utilize the Logistics Regression classifier for the accuracy of predicting the $s$ and using original $x$ to predict $y$. The to be dispelled s represents source dataset (i.e., domain) on DIGITS, and represents lighting condition on Extened YaleB, both are main-task *irrelevant* semantic variations.

Table.5.5.
Classification accuracy comparisons

| Method | Accuracy on Extended YaleB | | | |
|---|---|---|---|---|
| | $(y \mid d)$ | $(s \mid d)$ | $(y \mid l)$ | $(s \mid l)$ |
| Original $x$ as $d$ | 78.0% | 96.1% | - | - |
| Li | 82% | - | - | - |
| Louizos | 84.6% | 56.5% | - | - |
| Daniel | 86.6% | 47.9% | | - |
| Proposed | **90.1**% | **26.2**% | **8.7**% | **30.5**% |

Table.5.6.
Summary of the 40 face attributes provided with the CelebA and LFWA dataset. We expect the network learns to be invariant to the bolded and italicized attributes for our makeup face recognition task. *We noticed the degrades of recognition accuracy in CelebA dataset when dispelling these attributes.

| Att.Id | Attr.Def | Att.Id | Attr.Def | Att.Id | Attr.Def | Att.Id | | Attr.Def |
|---|---|---|---|---|---|---|---|---|
| 1 | 5'O Shadow | 11 | *Gray Hair** | 21 | Male | 31 | | *Sideburns* |
| 2 | Arched Eyebr | 12 | Big Lips | 22 | *Mouth Open* | 32 | | *Smiling* |
| 3 | Bushy Eyebr | 13 | Big Nose | 23 | *Mustache* | 33 | | *Straight Hair* |
| 4 | *Attractive* | 14 | Blurry | 24 | Narrow Eyes | 34 | | *Wavy Hair* |
| 5 | *Eyes Bags* | 15 | Chubby | 25 | *No Beard* | 35 | | *Earrings* |

| 6 | *Bald** | 16 | Double Chin | 26 | Oval Face | 36 | *Hat* |
| 7 | *Bangs* | 17 | *Eyeglasses* | 27 | *Pale Skin* | 37 | *Lipstick* |
| 8 | *Black Hair** | 18 | *Goatee* | 28 | Pointy Nose | 38 | Necklace |
| 9 | *Blond Hair** | 19 | *Makeup* | 29 | Hairline | 39 | Necktie |
| 10 | *Brown Hair** | 20 | Cheekbones | 30 | *Rosy Cheeks* | 40 | *Young** |

Table 5.7.
Comparisons of the rank-1 accuracy and TPR@FPR=0.1% on three makeup datasets.

| Dataset | PR2017 | | TCSVT2014 | | FAM | |
|---|---|---|---|---|---|---|
| | Methods | Acc \| TPR | Methods | Acc \| TPR | Methods | Acc \| TPR |
| | [Sharmanska, et al., 2012] | 68.0% \| - | [Louizos, et al., 2015] | 82.4% \| - | [Hu, et al., 2013] | 62.4% \| - |
| | [LeCun, et al., 1998] | 92.3% \| 38.9% | [LeCun, et al., 1998] | 94.8% \| 65.9% | [Kushwaha, et al., 2018] | 82.6% \| - |
| | VGG | 82.7% \| 34.7% | VGG | 84.5% \| 59.5% | VGG | 80.8% \| 48.3% |
| | Proposed | **94.6%** \| **45.9%** | Proposed | **96.2%** \| **71.4%** | Proposed | **91.4%** \| **58.6%** |

Table 5.8.
Face recognition accuracy on CelebA dataset

| Methods | Rank-1 accuracy |
|---|---|
| VGG | 85.4% |
| 19-head (1ID+18attr) | 81.1% |
| FLF | **92.7%** (↑22.7%) |

Table 5.9.
Face attribute recognition accuracy on CelebA and LFWA dataset. Two datasets are trained and tested separately.

| Methods | backbone | CelebA | LFWA |
|---|---|---|---|
| [Liu, et al., 2018] | AlexNet | 87.30% | 83.85% |
| [Louizos, et al., 2015] | VGG-16 | 91.20% | - |
| [Liu, et al., 2018] | InceptionResNet | 87.82% | 83.16% |
| [He, et al., 2018] | ResNet50 | 91.81% | 85.28% |
| FLF | VGG-16 | **93.26%** | **87.82%** |

[Liu, et al., 2019] evaluate the desired makeup-invariance property of the learned discriminative representation on three makeup benchmarks. To be detailed, [Liu, et al., 2019] train their framework using CelebA dataset [Liu, et al., 2018] which is a face dataset with 202,599 face images from more than 10K subjects, with 40 different attribute labels where each label is a binary value. [Liu, et al., 2019] adapt the $E_d$ and $C_d$ from VGG-16 [Perarnau, et al., 2016], and the extracted $d$ in testing stage are directly utilized for the open-set recognitions [Liu, et al., 2017], without fine-tuning on the makeup datasets as the VGG baseline method.

PR 2017 Dataset [Sharmanska, et al., 2012] collected 406 makeup and non-makeup images from the Internet of 203 females. TCSVT 2014 dataset [Guo, et al., 2013] incorporate 1002 face images. FAM dataset [Hu, et al., 2013] involves 222 males and 297 females, with 1038 images belonging to 519 subjects in total. It is worth noticing that all these images are acquired under uncontrolled condition. [Liu, et al., 2019] follow the protocol provided in [LeCun, et al., 1998], and the rank-1 average accuracy of FLF and state-of-the-art methods are reported in Table 6. as quantitative evaluation. The performance of [LeCun, et al., 1998], VGG-baseline and FLF are benefited from the large scale training dataset in CelebA. Note that the CelebA used in FLF and baseline, and even larger MS-Celeb-1M databases [Guo, et al., 2016] used in [LeCun, et al., 1998] have incorporated several makeup variations.

With the prior information about the makeup recognition datasets, [Liu, et al., 2019] systematically enforce the network to be invariant to the makeup-related attributes, which incorporate both the id-relevant variations (e.g., hair color) and id-irrelevant variations (e.g., smiling/not). Dispelling these id-relevant attributes usually degrades the recognition accuracy in original CelebA dataset, but achieve better generalization ability on makeup face recognition datasets.

Since these attributes are very likely to be changed for the subjects in makeup face recognition datasets, the FLF can extracts more discriminative feature for better generalization ability.

By utilizing the valuable side labels (both main-task and attributes) in CelebA in a controllable way, [Liu, et al., 2019] achieve more than 10% improvement over the baseline, and outperforms STOA by $\geq$5.5% $w.r.t$ TPR@FPR=0.1% in all datasets.

[Liu, et al., 2019] also take the open-set identification experiments in CelebA with an ID-independent 5-fold protocol. In Table 5, [Liu, et al., 2019] have shown which 18 attributes can increase the generalization in CelebA, while 6 attributes will degrade the accuracy in CelebA while improving the performance in Makeup face recognition. The accuracy of FLF on CelebA after dispelled 18 kinds of attributes is significantly better than its baselines. The VGG does not utilize the attribute label, the 19-head is a typical multi-task learning framework which can be distracted by task-$irrelevant\ s$.

Inversely, [Liu, et al., 2019] can flexibly change the main-task as attribute recognition and dispel the identity information. As shown in Table 8, FLF outperforms the previous methods with a relatively simple backbone following the standard evaluation protocol of CelebA and LFWA [Liu, et al., 2018] benchmarks.

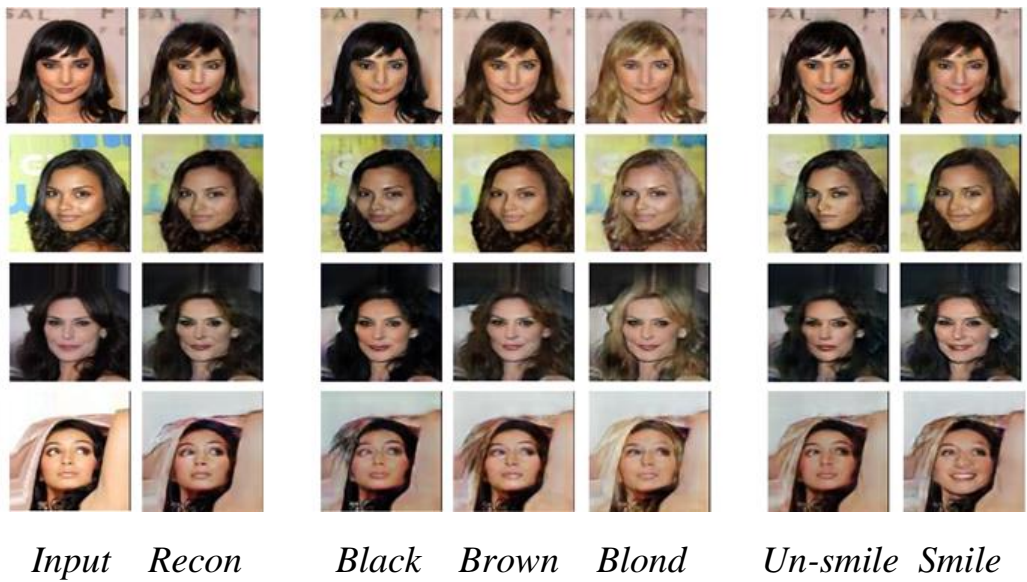

*Input   Recon      Black   Brown   Blond     Un-smile  Smile*

Figure 5.8. t-SNE visualization of images in Extended YaleB. The original images (b) are clustered according to their lighting environments, while the discriminative representation learned by our framework (a) is more likely to cluster with only identities.

To further verify the quality of semantic variations dispelling, [Liu, et al., 2019] acquire some of the conditional generated images in Fig.5.8, given the input samples from the test set of CelebA. Without changing attributes vector $s$, these three complementary parts maintain the most of information to reconstruct the input samples. Benefited from the information encoded in the latent variation vector $l$, the background can be well maintained. [Liu, et al., 2019] are able to change any semantic attributes incorporated in $s$, while keeping the $d$ and $l$ for identity-preserved attributes transform, which archives higher naturalness than previous pixel space IcGAN [Peng, et al., 2017]. The methods commonly used in vision to assess the visual quality of the generated images (*e.g.*, Markovian discriminator [Jayaraman, et al., 2014]) could totally be applied on top of their model for better texture, despite we do not focus on that.

The 5 hours training takes on a K40 GPU is 3× faster than pixel-level IcGAN [Peng, et al., 2017], without the subsequent training using the generated image for recognition and the inference time in the testing phase is the same as VGG.

The disgust face in the wild (DFW) dataset [Lample, et al., 2017] is a recently leased benchmark, which has 11157 images from 1000 subjects. The mainstream methods usually choose CelebA as pre-training dataset, despite it has a slightly larger domain gap with CelebA than these makeup datasets. In Table 9, [Liu, et al., 2019] show the FLF can largely improve the VGG baseline by 18% and 20.9% $w.r.t$ GAR@1%FAR and GAR@0.1%FAR respectively. It can also be used as a pre-training scheme (FLF+MIRA) to complementary with the state-of-the-art methods for better performance.

Table 5.10.
Face recognition on DFW dataset

| Methods | @1%FAR | @0.1%FAR |
| --- | --- | --- |
| VGG | 33.76% | 17.73% |
| FLF | 51.78% | 38.64% |
| | (↑18.02%) | (↑20.91%) |
| MIRA | 89.04% | 75.08% |
| FLF+MIRA | **91.30%** | **78.55%** |
| | (↑2.26%) | (↑3.47%) |

## 5.7. Discussion
### 5.7.1. Independent analysis

The three complementary parts are expected to uncorrelated to each other. The $s$ is marginally independent to the $d$ and $s$, since its short code cannot incorporate the other information. [Liu, et al., 2019] learn the $d$ to be discriminative to the main recognition task and marginally independent to $s$ by maximizing the certainty of making main task predictions and uncertainty of inferring the semantic variations given the $d$. Given the $l$, minimizing the certainty of making main task ($y$) predictions can makes $l$ marginally independent to the $d$ and some of the task-dependent $s$.

Considering the complexity of the framework, [Liu, et al., 2019] do not strictly require the learned $l$ to be marginally independent to task-irrelevant $s$. The ground truth label of $l$ also does not exist in the datasets to supervise the $d$ to be marginally independent to latent variation $l$. Instead, [Liu, et al., 2019] limit the output dimension of $E_d$ and $E_l$ as an information bottleneck to implicitly require $d$ and $l$ incorporate little unexpected information [Kingma, et al., 2014, Theis, et al., 2015]. Additionally, a reconstruction loss is utilized as the complementary constraint, which avoids the $d$ and $l$ containing nothing.

### 5.7.2. Equilibrium condition

Several trade-off parameters constrained between 0 and 1 are used to balance the judiciously selected loss functions. The $E_l$ is trained to minimize the $(-\mathcal{L}_{C_l} + \lambda \mathcal{L}_{rec})$, where the $\lambda$ is used to weight the relevance of the latent representation with the class label, and the quality of reconstruction.

The $E_d$ is updated by minimizing the $(\mathcal{L}_{Dis} - \alpha \mathcal{L}_{Dis} + \beta \mathcal{L}_{rec})$. The $\mathcal{L}_{rec}$ works as a complementary constraint, the $\beta$ is usually given a relatively small value. [Liu, et al., 2019] omit this term for simplicity to analyze the function of $\alpha$. The objective of semantic variation dispelling can be formulated as:

$$\min_{E_d, C_d} \max_{Dis} \mathbb{E}_{x,s,y \sim q(x,s,y)} \left[ -\log p_{C_d}(y \mid E_d(x)) + \alpha \log p_{Dis}(s \mid E_d(x)) \right] \quad (5.18)$$

[Liu, et al., 2019] explain how the task of $d$ preserving and $s$ eliminating are balanced in the game under non-parametric assumptions (*i.e.*, assume a model with infinite capacity). Two scenarios are discussed where $s$ is dependent/independent to $y$.

Considering that both the $E_d$ and $C_d$ use $d$ which is transformed deterministically from $x$, [Liu, et al., 2019] substitute $x$ with $d$ and define a joint distribution $\tilde{q}(d,s,y) = \int_x \tilde{q}(d,x,s,y) dx = \int_x q(x,s,y) p_{E_d}(d \mid x) dx$. Since the $E_d$ is a deterministic transformation and thus the $p_{E_d}(d \mid x)$ is merely a delta function denoted by $\delta(\cdot)$. Then $\tilde{q}(d,s,y) = \int_x q(x,s,y)\delta(E_d(x) = d)dx$, which depends on the transformation defined by the $E_d$. Intuitively, $d$ absorbs the randomness in $x$ and has an implicit distribution of its own. [Liu, et al., 2019] equivalently rewrite the Eq. 18 as:

$$\min_{E_d, C_d} \max_{Dis} \mathbb{E}_{d,s,y \sim \tilde{q}(d,s,y)}\left[-\log p_{C_d}(y \mid d) + \alpha \log p_{Dis}(s \mid d)\right] \quad (5.19)$$

To analyze the equilibrium condition of the new objective Eq. 19, [Liu, et al., 2019] first deduce the optimal $C_d$ and $Dis$ for a given $E_d$, then prove its global optimality.

For a given fixed $E_d$, the optimal $C_d$ outputs $p^*_{C_d}(y \mid d) = \tilde{q}(y \mid d)$, and the optimal $Dis$ corresponds to $p^*_{Dis}(s \mid d) = \tilde{q}(s \mid d)$. [Liu, et al., 2019] use the fact that the objective is functionally convex *w.r.t.* each distribution, and by taking the variations, we obtain the stationary point for $p_{C_d}$ and $p_{Dis}$ as a function of $\tilde{q}(d,s,y)$. The optimal $p^*_{C_d}(y \mid d)$ and $p^*_{Dis}(s \mid d)$ given in *Claim 1* are both functions of the encoder $E_d$. Thus, by plugging $p^*_{C_d}$ and $p^*_{Dis}$ into the Eq. 19, it can be a minimization problem only *w.r.t.* the $E_d$ with the following form:

$$\min_{E_d} \mathbb{E}_{\tilde{q}(d,s,y)}[-\log \tilde{q}(y \mid d) + \alpha \log \tilde{q}(s \mid d)] = \min_{E_d} H(\tilde{q}(y \mid d)) - \alpha H(\tilde{q}(s \mid d)) \quad (5.20)$$

where the $H(\tilde{q}(y \mid d))$ and $H(\tilde{q}(s \mid d))$ are the conditional entropy of the distribution $\tilde{q}(y \mid d)$ and $\tilde{q}(s \mid d)$ respectively.

As we can see, the objective consists of two conditional entropies with different signs. Minimizing the first term leads to increasing the certainty of predicting $y$ based on $d$. In contrast, minimizing the second term with the negative sign amounts to maximizing the uncertainty of inferring $s$ based on $d$, which is essentially filtering out any information about semantic variations from the discriminative representation.

• For the cases where the attribute $s$ is entirely independent of main recognition task, these two terms can reach the optimum simultaneously, leading to a win-win equilibrium. For instance, with the lighting effects on a face image removed, we can better identify the subject. With sufficient model capacity, the optimal equilibrium solution would be the same regardless of the value of $\alpha$.

• We may also encounter cases where the two objectives compete with each other. For example, learning a task-dependent semantic variation dispelled representation may harm the original main-task performance. Hence the optimality of these two entropies cannot be achieved at the same time and the relative strengths of the two objectives in the final equilibrium are controlled by $\alpha$.

### 5.8. Conclusion

How to extract a feature representation that not only be informative to the main recognition task, but also irrelevant to some specific notorious factors is an important objective in visual recognition. This chapter systematically summarized the possible factors and introduced two practical solutions to achieve the disentanglement in a controllable manner. Specifically, [Liu, et al., 2017] derive the (*N+M*)-tuplet clusters loss and combine it with softmax loss in a unified two-branch FC layer joint metric learning CNN architecture to alleviate the attribute variations introduced by different identities on FER. The efficient identity-aware negative-mining and online positive-mining scheme are employed. After evaluating performance on the posed and spontaneous FER dataset, [Liu, et al., 2017] show that the proposed method outperforms the previous softmax loss-based deep learning approaches in its ability to extract expression-related features. More appealing, the ($N+M$)-tuplet clusters loss function has clear intuition and geometric interpretation for generic applications. [Liu, et al., 2019] present a solution to extract discriminative representation inheriting the desired invariance in a controllable way, without the paired semantic transform example and latent labels. Its recognition does not need generated image as training data. As a result, [Liu, et al., 2019] show that the invariant representation is learned, and the three parts are complementary to each other. Considering both the labeled semantic variation and the unlabeled latent variation can be a promising developing direction for many real-world applications.

176

of the 2015 ACM on International Conference on Multimodal Interaction, 2015, P.491-496